%% file: root.tex
\documentclass[letterpaper, 10 pt, conference]{ieeeconf}  
\usepackage{graphicx}
\usepackage{hyperref}

\IEEEoverridecommandlockouts                              

\title{\LARGE \bf
 Additively Manufactured Open-Source Quadruped Robots for Multi-Robot SLAM Applications
}

\author{Zachary Fuge$^{1}$, Benjamin Beiter$^{2}$ and Alexander Leonessa$^{3}$
\thanks{$^{1}$Zachary Fuge is a student with the Department of Mechanical Engineering,
        Virginia Tech,635 Prices Fork Rd, Blacksburg, VA 24060
        {\tt\small zfuge@vt.edu}}%
\thanks{$^{2}$Benjamin Beiter is with the Department of Mechanical Engineering, 
        Virginia Tech,635 Prices Fork Rd, Blacksburg, VA 24060}
\thanks{$^{3}$Alexander Leonessa is with Faculty of Mechanical Engineering,
        Virginia Tech,635 Prices Fork Rd, Blacksburg, VA 24060}
}

\begin{document}

\maketitle
\thispagestyle{empty}
\pagestyle{empty}

\begin{abstract}


This work presents the design and development of the quadruped robot ``Squeaky" to be used as a research and learning platform for single and multi-SLAM robotics, computer vision, and reinforcement learning. Affordable robots are becoming necessary when expanding from single to multi-robot applications, as the cost can increase exponentially as fleet size increases. SLAM is essential for a robot to perceive and localize within its environment to perform applications such as cave exploration, disaster assistance, and remote inspection. For improved efficiency, a fleet of robots can be employed to combine maps for multi-robot SLAM. Squeaky is an affordable quadrupedal robot, designed to have easily adaptable hardware and software, capable of creating a merged map under a shared network from multiple robots, and available open-source for the benefit of the research community.

\end{abstract}


\input{sections/1_Introduction}

\input{sections/2_OverallDesign}

\input{sections/3_SLAM}
\input{sections/4_Results}
\input{sections/5_Conclusion}

\bibliographystyle{unsrt}
\bibliography{TREC_SLAM_new}

\end{document}

%% file: sections/1_Introduction.tex
%

\section{Introduction}

There have been many improvements in SLAM (Simultaneous Localization and Mapping) algorithms in recent years. These can be classified into two primary approaches for 2D Lidar SLAM, which are filter-based and graph-based. These implementations have been integrated into the ROS framework as packages such as GMapping \cite{grisetti_improving_2005}, KartoSLAM \cite{KartoSLam}, Cartographer \cite{cartographer}, HectorSLAM \cite{HectorSLAM2011} and more recently SLAM toolbox \cite{macenski_slam_2021}. With these improvements, SLAM is becoming significantly more reliable and capable, and integration of multi-robot SLAM applications is becoming possible. The most recent approach for multi-robotic SLAM is Kimera-Multi, this approach demonstrates a distributed system, where each robot is mapping and able to merge each others' map in peer-to-peer communication \cite{chang_KimeraMulti_2021}. In simple terms, SLAM is an algorithm a robot uses to both map and localize itself within its surroundings as it moves through them, either autonomously or via remote control. This allows a robot to perceive the world and is helpful in many applications, including warehouse navigation and industrial applications \cite{keith2024reviewautonomousmobilerobots}, \cite{warehouse2021}. These algorithms require reliable mobile robots that can navigate environments successfully, and if the robot must operate in human areas and rough terrains, the current best option is legged robots. For research or education groups, access to such robots can be a barrier to implementing SLAM-capable devices in real and custom environments. To improve access to affordable SLAM-capable devices for human and outdoor navigation, this paper presents the design and construction of an open-source, small-scale, additively manufactured (AM) quadruped robot. This robot is equipped with lidar and depth vision sensing to enable mapping and a communication framework that connects multiple robots for simultaneous, cooperative SLAM. This quadruped, Squeaky, is affordable and open-source\footnote{This robot is open-sourced and available at the TREC gitlab: \href{https://gitlab.com/trec-lab}{gitlab.com/trec-lab}.}, can be printed on most commercially available desktop 3D printers, and hosts a system capable of large-scale, multi-robot SLAM, only constrained by connectivity to a shared network. 


\begin{figure*}[t]
\centering
\includegraphics[width=\linewidth]{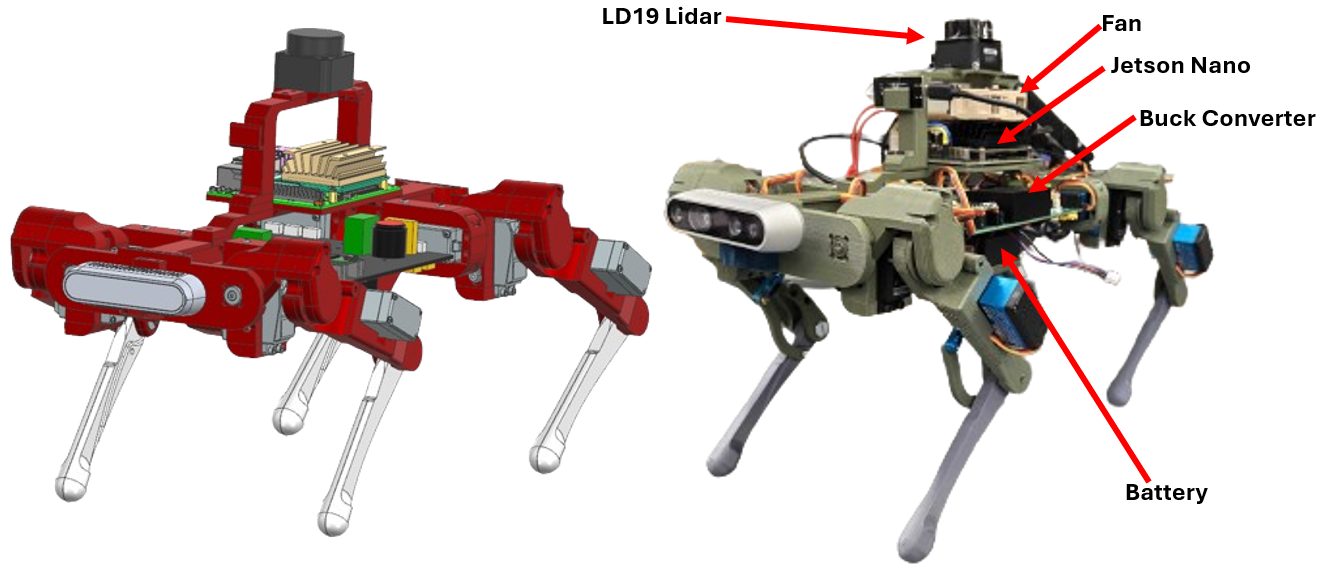}
\caption{Images of the Robot, CAD and Physical Version. Robot can be found \href{https://gitlab.com/trec-lab}{gitlab.com/trec-lab}}	
\label{fig:CadAndPhysical}
\vspace{-0.5 cm}
\end{figure*}

There have been previous full and partial AM quadrupeds that target affordability or ease of assembly \cite{kim_design_2021}, \cite{kau_Stanford_2021}. The benefit of AM is that it is cheaper, easy to build, redesign, or replace parts, and accessible to anyone with a 3D printer. Hence, with 3D printers, there has been an influx of quadrupeds aiming for affordability or capability. When going for affordability, they aim to remove the machining cost by using 3D-printed parts for the structural components. The primary cost of robots is the motors and other electrical parts. This can be seen with PADWQ, a 3D-printed robot with powerful motors \cite{kim_design_2021} and a base cost of \$7123 and with perception it is \$7692. Even with its increased affordability compared to other robots of it size, the price may still be a concern to some research groups wanting to test SLAM capabilities on legged robots. Another open-source quadruped includes the Stanford Pupper \cite{kau_Stanford_2021}. This quadruped is highly capable of trotting and jumping and is smaller in size and weight compared to PADWQ. The cost is better compared to PADWQ at \$892 for the base version, but it could still be considered unaffordable to some research groups if they want to expand into multi-robot SLAM. Considering the cost, RealAnt is a great choice at \$354 for the base cost for the robot but has been developed to experiment with reinforcement learning algorithms, not SLAM. Hence, this robot is designed to be connected to external power and computer when testing \cite{boney_realant_2022}. Charlotte is a quadruped designed for exploration and SLAM applications  \cite{garcia-cardenas_charlotte_2020}. It base robot cost is \$540 and has an arachnid-like chassis. As the chassis is larger, it may be unable to map different areas due to this size constraint. Overall, the gap that remains is an affordable and capable robot that can be expanded into multi-robotic applications.



To address this gap, we present a low-cost, open source, 3D-printed robot capable of SLAM. The specific contributions are the AM part design built off the Squeaky platform, custom-designed electronics for the whole robot, and a simple communication and integration framework for multi-SLAM robots. These improvements allow for an easy-to-manufacture, affordable, mobile, and SLAM-capable robot. This robot is designed to be a modular robot that can allow for sensor and parts to be easily changed. The software framework also allows for ease of implementation of different algorithms based on research goals, and enables easy setup and execution of multi-robotic SLAM, as shown in the Results. Through its camera, more computer vision or visual-SLAM algorithms can be investigated. Overall, this robot is an affordable, accessible platform for implementing and testing SLAM algorithms for research, education, and small-scale mobile robot applications. 


The rest of the paper will be structured as follows. Section \ref{sec:overall_design} presents Squeaky's design modified for SLAM, including mechanical, electrical, and the overall system design framework. Section \ref{sec:slam} will present the SLAM implementation on the robot, including capabilities and limitations, compatible algorithms, and multi-robot implementation. Section \ref{sec:results} will present results validating the robot's functionality as a capable SLAM platform with various algorithms and multiple robots. Section \ref{sec:conclusion} will conclude the paper.

%% file: sections/2_OverallDesign.tex

\section{Overall Design} \label{sec:overall_design}

Presented within this paper is the complete design of a fully 3D-printed quadruped that can be used for multi-purpose tasks such as SLAM, navigation, and legged locomotion. The robot's inexpensive and easy-to-manufacture nature lends itself to multi-robot applications. The robot will be broken down into multiple components in the mechanical design, electrical design, and system overview. 

Building upon the initial open-source Squeaky design \cite{schoedel_development_2023}, this paper presents an updated quadruped with a focus on having a more powerful microprocessor, a larger chassis, a custom PCB, and new sensors. The system framework has also been changed to accommodate the perception sensors and for ability to configure the controller within the ROS framework. This had been set up to allow for faster optimization since this robot has multiple research avenues that can be taken. A key point of this robot is modularity, hence the design being largely 3D printed and the system framework allowing interchangeability.

The Squeaky presented here includes an Nvidia Jetson NANO (NANO) as the onboard computer because of the small size and powerful performance that it offers. This allows for more computationally heavy algorithms to be performed on the robot itself. The microprocessor operating system on this robot is Ubuntu 18.04. The Robot Operating System (ROS) Framework is Melodic and utilized for easier integration of sensors, nodes, and messages between these nodes. 

%


\subsection{Mechanical Design}

The mechanical design of squeaky can be broken down into two larger sections: the legs and the chassis. Since the legs design has been previously demonstrated in \cite{schoedel_development_2023} with no significant changes made in this paper, the focus will be the upgrades to the chassis design.

The chassis's changes are the widening and lengthening of the main body. Previously, it was 275 mm and has now been lengthened to 350 mm to accommodate a NANO, a larger buck converter, and a custom control board. These are all placed on the custom-design PCB control board(further discussed in the electrical design section). Overall, the placement of components affects the design of the PCB as the pinouts for the NANO and buck converter need to be connected to the custom PCB via external wires. 

\begin{figure}[!t]
\centering
\includegraphics[width=\linewidth]{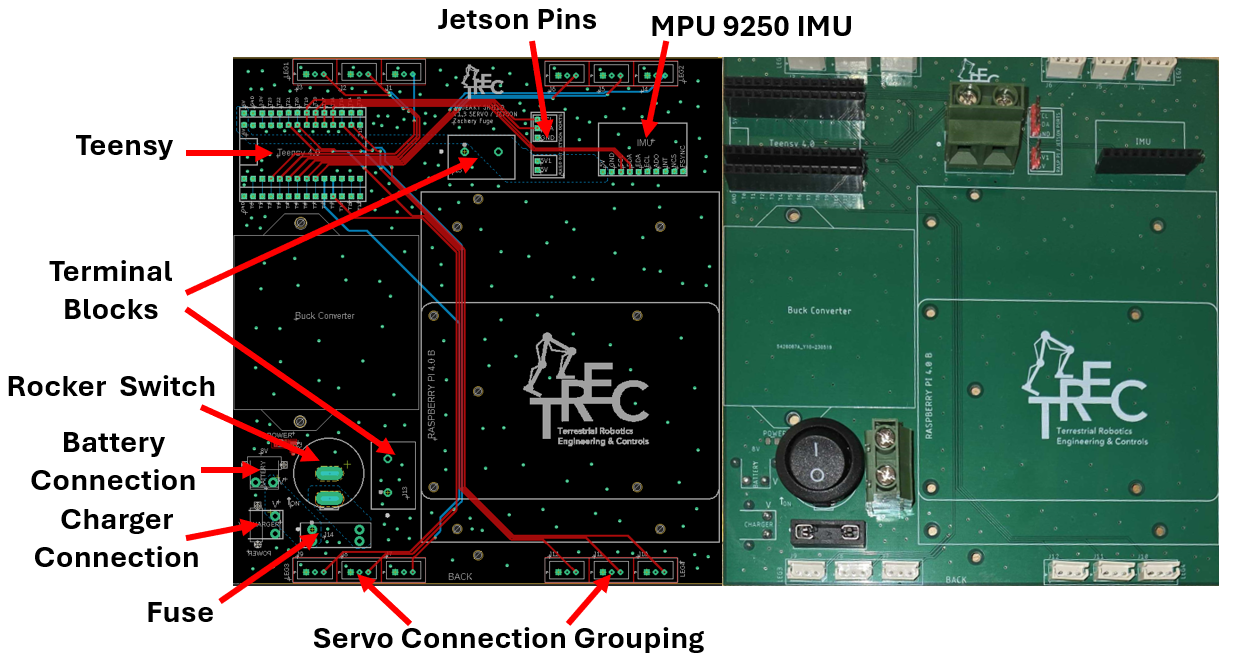}
\caption{Left image is PCB in CAD and right image is Physical PCB}	
\label{fig:PCB}
\vspace{-0.7 cm}
\end{figure}

This perception variant of Squeaky includes a design for holding the NANO which can be seen in Figure \ref{fig:CadAndPhysical} centered on the entire squeaky chassis. Early designs had the NANO placed on the PCB, but had to be shifted up due to wire routing for the buck converter. Placement of components on the PCB can be seen within \ref{fig:PCB}. With the current NANO placement, it still has space to connect to the GPIO's on the custom PCB for control of the robot. Then with the AM parts, a large bracket was designed and implemented to mount the 2D LD19 LDROBOT lidar above the robot chassis. It was shifted higher due to a fan that was used to handle cooling of the NANO microprocessor. The fan was added  since the microprocessor runs 20-21C on idle, but when running computationally heavy algorithms, it will spike to 39C+ and causes a thermal shutdown. After the addition of the fan, the temperature maintains 31-32C when more intensive algorithms such as SLAM are running. Part placement can be seen in Figure \ref{fig:CadAndPhysical}.

The attachment for the lidar is made of two separate pieces, the bracket itself an interface piece that connects the lidar to the bracket. This was done intentionally to simplify the 3D printing of these components by removing the need for excessive supporting structure when printing the components and shortening the overall print time. The bracket also includes an outcropping on one column to support the lidar control board that acts as an interconnected piece between theoNANO USB port and the 2D Ld19 lidar.


Squeaky is designed to be modular in its ability to add and remove components as necessary. This can be seen in the front and back of the main chassis as there sits a T column on the front and back. This is meant for attachments to be created and be able to slide and lock in place in this area. These are currently utilize to hold the D435i camera on the front and on the back is two antennas for WIFI and Bluetooth. Lastly the D435i camera has a 1/4-20 UNC thread mounting point which is utilized to attach the camera to the part that locks into the T-column.

\subsection{Electrical Design}

The electrical design of this robot is straightforward. There are 12 Servo motors that handle the locomotion. These are all connected to a custom design PCB that serves as the center of the quadruped. The servo grouping can be seen on Figure \ref{fig:PCB} labelled as Servo Connection Grouping. Each Grouping represents one leg such as the shoulder, thigh and shin servoes. The battery is placed under the board and fitted with a 3d printed bracket to keep it in place. This placement allows the battery to be plugged into the bottom of the PCB for easy power access. Because the battery is always connected to the PCB once placed, a fuse is added between the rocker switch and the battery. This allows for the fuse to protect the board if needed, as well as serving as a cut-off denying power to the rest of the robot in the case of an accidental bump of the rocker switch, such as when transporting it. 

\begin{figure}[!t]
\centering
\includegraphics[width=0.9\linewidth]{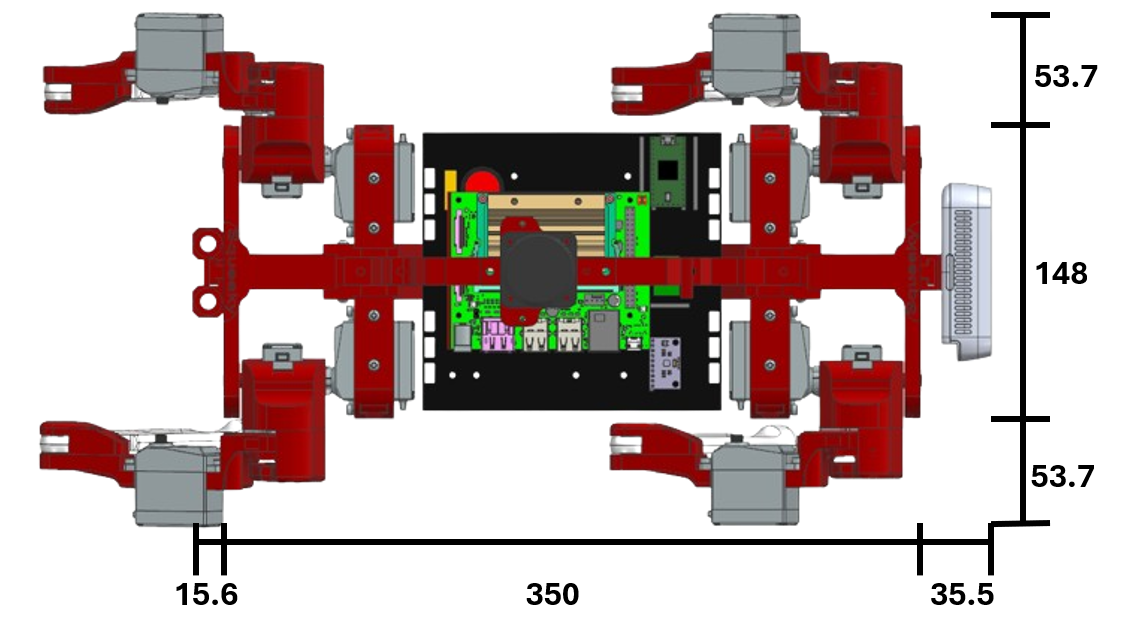}
\caption{Squeaky Dimension, All dimensions are in mm}	
\label{fig:SqueakyDim}
\vspace{-0.7 cm}
\end{figure} 

As previously stated, The battery is connected to a simple circuit on the PCB, allowing for power to pass through a fuse and then a rocker switch to turn on/off the robot. There is also another port to connect a charger to the battery and charge through the PCB on the other side of the fuse, so the fuse can be removed and still be able to charge the battery. The battery also has a wire that is connected to the cells and must be connected to the charger.

To control the servo, the pwm pins are connected to individual GPIO pins on a Teensy 4.0 which controls the locomotion of the entire robot. The custom PCB has a place where the Teensy has been designed to sit in. The Teensy receives commands over I2C from the NANO, which is the overall brain of the robot. The NANO is also powered through the same battery as the rest of the robot, as there are 5 pins on the custom PCB, 2 for I2C (SCL and SDA), 2 for power, and 1 for ground. 

To handle power distribution on the PCB, the PCB has 4 layers. The top layer is the signal, 2nd layer is the ground plane, 3rd layer is a power plane, and 4th layer is the signal. Extra space is connected to the ground plane on the two signal layers in order to eliminate possible noise between signal lines. The power plane is connected directly to the battery after the rocker switch. In this case, a 7.4 V Lipo battery powers the servo motors. The buck converter transforms that to a 5V output for the logic circuits. A sectioned off copper sheet is placed on the bottom signal layer which is a 5V layer and powers all associated components that require this power level. 


In the previous squeaky version \cite{schoedel_development_2023}, a smaller buck converter only outputted a low current. This was unsuitable for this new variant design with high power demands primarily because of the NANO and its peripherals. This older buck converter was replaced with a 5V 10A output buck converter. This has higher specifications than what is strictly required since the robot only pulls around 6.5 Amp. 6 Amp is the max of what the NANO will pull if all peripherals and a computationally heavy algorithm are running. The remaining 4 Amp is reserved for the remaining logic circuits. 

This chosen buck converter could not be soldered to the board and only had wire outputs. The PCB has two pairs of terminal blocks whose specifications meet the current needs of the logic power of Squeaky. The wire gauge is 12 AWG and has been increased to ensure no limits are reached when powering. 



\begin{figure}[!t]
\centering
\includegraphics[width=\linewidth]{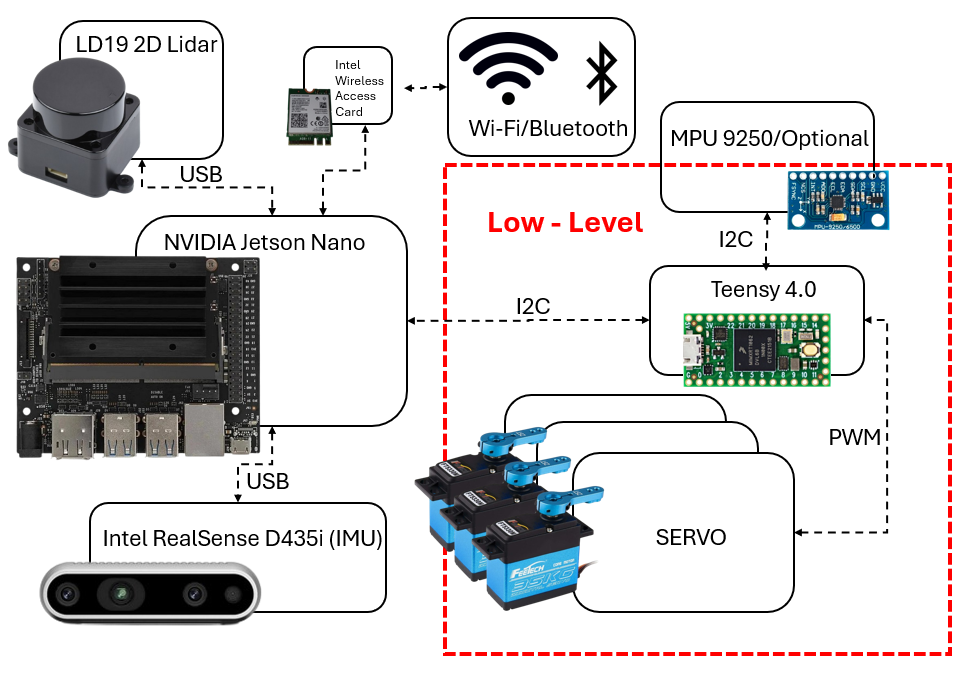}
\caption{Overview of the system components for communication and control of Squeaky}	
\label{fig:experimentImage}
\vspace{-0.7 cm}
\end{figure}




\subsection{System Overview}

The System Framework, summarized in Figure \ref{fig:experimentImage}, can be broken into two systems, a Low-Level (LL) system and a High-Level (HL) system. Each system has different tasks and priorities.

The LL system is the Teensy 4.0 microcontroller, with the main task of controlling the 180-degree servo motors. Since these servos have a range of motion (ROM) of only 180 degrees, they must be calibrated by finding the zero value according to the servo and placing the servo horn in the correct position. As servos are absolute in their position even though they are open-loop and provide no feedback, if calibrated in a known position, then the legs can be zeroed out and set to the desired position through the Teensy. An offset variable is also created within the LL to calibrate the servos into the final desired position. This is all done for the ROS controller in the HL system, which contains a model of the robot, and is where the High-level control is performed. The I2C line between the Teensy and the NANO is the connecting bridge between the HL and LL. All the large computations needed are performed in the HL such as calculations for the desired angles that the servo motors must rotate to. The desired angles of the motors are transmitted from the HL to LL, which is currently the only transmission between the two systems. 

\begin{figure}[!t]
\centering
\includegraphics[width=\linewidth]{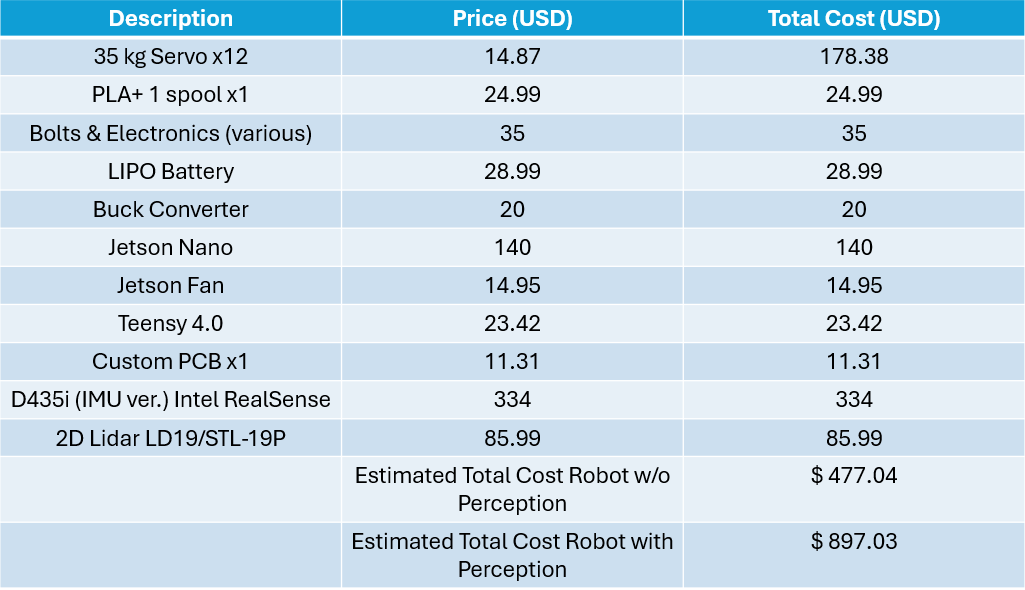}
\caption{Bill of Material of Squeaky with and without perception}	
\label{fig:cost}
\vspace{-0.7 cm}
\end{figure}

\begin{figure}[!b]
\vspace{-0.7 cm}
\centering
\includegraphics[width=\linewidth]{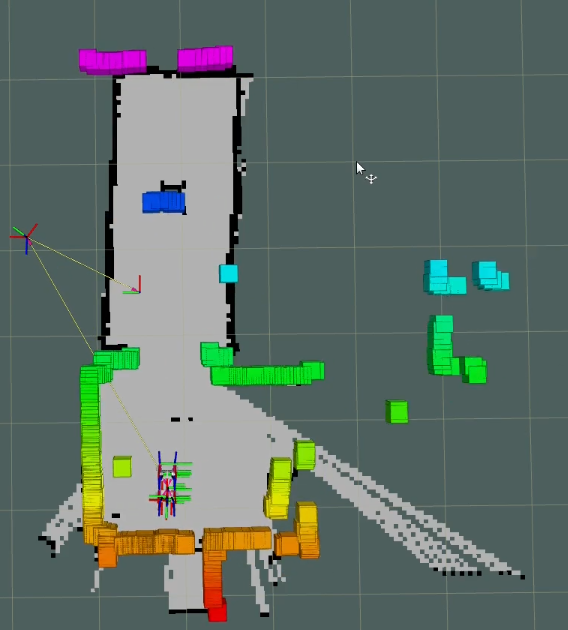}
\caption{SLAM Map Generated with Gmapping}	
\label{fig:Gmapping}
\vspace{-0.7 cm}
\end{figure}

\begin{figure*}[!t]
\centering
\includegraphics[width=\linewidth]{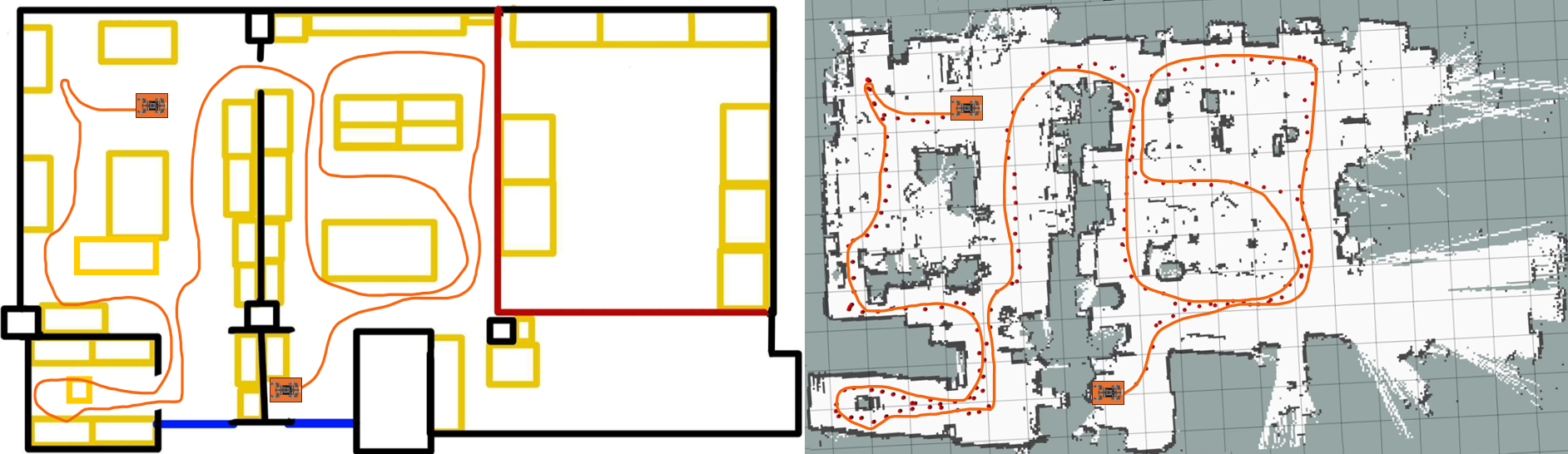}
\caption{Left image shows simplify building schematic of lab space with yellow representing approximate furniture placement. The Orange line is squeaky pathing which can be compared with the Actual map that squeaky created.}	
\label{fig:SlamToolboxSingle}
\vspace{-0.6 cm}
\end{figure*}


The HL system performs various other critical task for experimentation. Firstly, perception sensors are connected to the USB on the NANO to allow the robot to visualize its surroundings. In this case, the primary sensor is the 2D Innomaker LD19 lidar, which sits at the very top of the robot. Another sensor includes the D435i Intel Realsense, which is currently only used for its IMU data but, in the future, will be used for its depth camera and RGB camera, which is one direction of improvement that could be pursued for this robot, implementing visualization type algorithms for identification and tracking.




Currently on startup, four ROS launch file may be utilized, in this case the first two launch files are critical for overall robot control while the other two are optional depending on what the application is. The first launch file is for the d435i camera node, LD19 lidar, and I2C transmitter node to the LL. The second launch file contains the overall robot system, which utilizes the CHVMP framework, an open-source GitHub for ROS robots in quadruped control.\footnote{This framework is open-sourced and available on github: \href{https://github.com/chvmp/champ}{github.com/chvmp/champ}.} This launch will include many algorithms, such as a ROS controller configuration and state estimation for the robot. 

The third node to launch is a Squeaky's teleoperation node that takes in a joystick command and outputs a command velocity to control the robot. Then the fourth node to launch, through this same package, GMapping will be started and a simple 2D mapping of the environment can be made. This allows all sensor configurations to be tested to confirm that the robot is operating as expected. Afterwards, it can be replaced with more sophisticated SLAM algorithms depending on the sensors that are included on the robot and what the operators goals are.



The IMU from the d435i is also filtered. First, it is received through a driver package by INTEL RealSense for ROS which outputs a raw acceleration and gyration frame. Then a madgwick filter node receives this raw data and outputs an orientation. This is used in the robotic odometry which is a requirement to utilize SLAM algorithms such as GMapping and SLAM toolbox.



\begin{figure*}[!t]
\centering
\includegraphics[width=\linewidth]{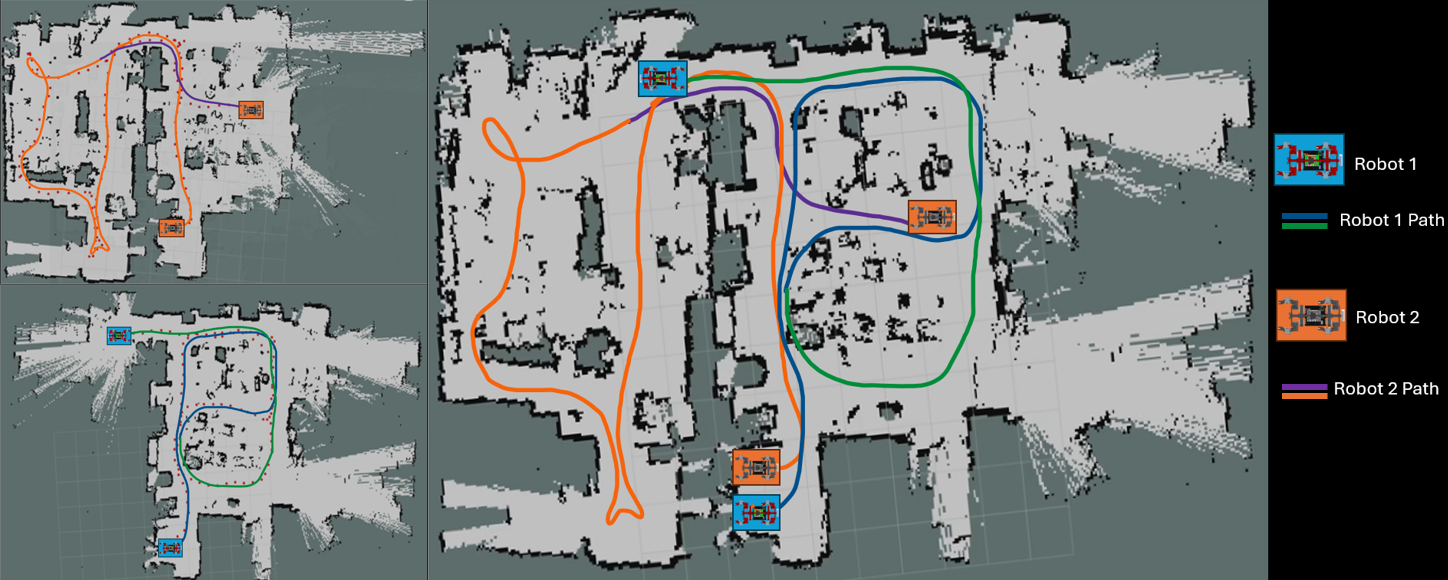}
\caption{Multi Robot SLAM Map Generated Using SLAM Toolbox and Map Merging. Top left is robot 1 map, Bottom left is robot 2 map, and right is merged map of 1 and 2.}	
\label{fig:SlamToolboxMulti}
\vspace{-0.6 cm}
\end{figure*}


%% file: sections/3_SLAM.tex
\section{SLAM} \label{sec:slam}


SLAM can be broken down into a variety of approaches depending on sensor configuration. In this case, the sensor used is a LDROBOT LD19 2D lidar for mapping and will focus on the two primary types of algorithms utilized for this type of data, which are Filter-based approaches and Graph-based approaches. Filter-based approaches treat SLAM as a state estimation problem and utilize filtering to handle changes of the robot pose over time and sensor input for mapping. Common filters used are the Extended Kalman filter \cite{EKF2004}, \cite{EKF2007} and Rao-Blackwellize particle filter \cite{grisetti_Improved_2007}, which is the basis for GMapping. Meanwhile, Graph-base approaches structure the SLAM problem as a nodal structure, with each node containing the robot's pose and sensor information at that position. Then, with the next node, a transformation must be performed between the two nodes, either based on sequential odometry data or by aligning robot observation at the prior and new node. Packages that follow a graph-base approach include Karto \cite{KartoSLam} and SLAM toolbox \cite{macenski_slam_2021} built off Open Karto.

This robot is versatile because any specific SLAM package can be implemented as long as it is compatible with the current sensor configuration. The specific package utilized for this paper is SLAM toolbox. This is a feature-rich package that includes asynchronous and synchronous mapping modes, as well as the ability to map an area, save it,load the saved pose-graph/map, and continue mapping. The list of features is extensive and is covered in \cite{macenski_slam_2021}. 
SLAM toolbox was used because the final map is published on a namespace/map topic, and with multiple Squeaky's under the same ROS master, a map merging node can be executed.\footnote{This merging node is open-sourced and available on github: \href{https://github.com/hrnr/m-explore/tree/melodic-devel?tab=readme-ov-file}{github.com/hrnr/m-explore/}.} This map merging node can work with n number of robots /maps and outputs one merged map. This is adjusted based on the mode being used. Since these robots are in a real environment with an unknown position, the feature-matching component of the algorithm is utilized. In this case, a large area must be mapped to merge the two robot individual maps into one.

%% file: sections/4_Results.tex
\section{Results} \label{sec:results}





To show affordability, the final cost of all components of the robot is summarized in Figure \ref{fig:cost}. In order to show the capabilities of the platform, two different SLAM packages through ROS are implemented. Firstly, GMapping is utilized, which is a very basic algorithm that utilizes Rao-Blackwellize particle filtering and scan matching to keep the robot localized. Demonstrating this algorithm's implementation, a simple map of a small office space is seen within Figure \ref{fig:Gmapping}. Gmapping, in this case, is set up to show the robot model and the points of what the lidar currently sees. The solid black represents occupied space, white represents empty, and grey/green represents unknown space. In this implementation, GMapping does not track the path of the robot over time. In this particular figure, the robot mapping is being recorded while the different points of the lidar are displayed. It does not record into its overall map until a confidence level is hit.

SLAM toolbox is the next package utilized to show the modularity of the high-level framework. Instead of launching Gmapping, the SLAM toolbox mapping algorithm is launched, which opens a large amount of capabilities as previously mentioned in Section \ref{sec:slam}. This algorithm is for lidar 2D mapping, but its expanded capabilities will allow for multi-robotic implementation. SLAM toolbox also tracks the odometry over time to see the robot moving over the environment when mapping and displays these markers as red dots. SLAM toolbox implementation is seen in Figure \ref{fig:SlamToolboxSingle}, showing the start point and the path. The path looks erratic, but this is due to the robot pathing over a similar trajectory within the environment to further refine its internal map. The accuracy of this map can be compared to the left image in Figure \ref{fig:SlamToolboxSingle} which shows a simplified floor plan of the lab space with furniture placed as a approximation of where it during the experiment. Pathing of the robot was taken from the actual SLAM map and placed over the actual map to give a easier visual comparison.

To show multi robotic implementation, Two robots are placed in the same room and same orientation, this is to allow them to build a similar map in the beginning as the m-explore\footnote{This merging node is open-sourced and available on github: \href{https://github.com/hrnr/m-explore/tree/melodic-devel?tab=readme-ov-file}{github.com/hrnr/m-explore/}.} package that is doing the merges is set to unknown positions and will merge these maps based upon similar features. This can be seen with the path being taken within the top of Figure \ref{fig:SlamToolboxMulti}. To help with visualization, a pathing was drawn out for each robot to see where it went over time. This is a simplification that is based on the red dots that are normally plotted for each robot as it navigates the environment and maps. Figure \ref{fig:SlamToolboxMulti} shows Squeaky 1 and 2 individual maps and then, on the right, the overall merged map. To further distinguish the exact path taken, a different color was utilized when crossing back over itself to show the direct path the robot took. As shown, it can be seen that multiple robots can be easily integrated into this current framework setup to perform cooperative SLAM. This furthers exemplifies the modularity of Squeaky and its potential in applications of research.

%% file: sections/5_Conclusion.tex
\section{Conclusion} \label{sec:conclusion}

In this work, we presented the overall design and implementation of affordable quadruped "Squeaky" and showed a implementation of multi-robotic SLAM. This paper discussed the mechanical design improvements made from previous work \cite{schoedel_development_2023}, as well as detailing the electrical design, and system framework for the entire robot. Further, within the results, both single robot SLAM and Multi SLAM were demonstrated, and multi slam showed two robots mapping and a final merged map was created. The open-source design provides researchers with an affordable option for pursuing quadrupedal and legged locomotion, SLAM, and collaborative robotics studies.


Future work includes expanding multi-SLAM capabilities, such as redistributing the combined map into each robot's primary map to continue mapping from there. Other research areas to be explored will be vision-related areas and visual-based SLAM.